# Open video data sharing in developmental and behavioural science

*Peter B Marschik*[1,2,3,4]~*  *Tomas Kulvicius*[1,5,]*, *Sarah Flügge*[5], *Claudius Widmann*[1], *Karin Nielsen-Saines*[6], *Martin Schulte-Rüther*[1,4], *Britta Hüning*[7], *Sven Bölte*[2,8,9], *Luise Poustka*[1,4], *Jeff Sigafoos*[10], *Florentin Wörgötter*[4,5], *Christa Einspieler*[3], *Dajie Zhang*[1,3,4]

1. Child and Adolescent Psychiatry and Psychotherapy, University Medical Center Göttingen, Göttingen, Germany
2. Center of Neurodevelopmental Disorders (KIND), Centre for Psychiatry Research; Department of Women's and Children's Health, Karolinska Institutet, Stockholm, Sweden
3. iDN – interdisciplinary Developmental Neuroscience, Division of Phoniatrics, Medical University of Graz, Graz, Austria
4. Leibniz-ScienceCampus Primate Cognition, Göttingen, Germany
5. Department for Computational Neuroscience, Third Institute of Physics-Biophysics, Georg-August-University of Göttingen, Göttingen, Germany
6. Division of Pediatric Infectious Diseases, David Geffen UCLA School of Medicine, Los Angeles, USA
7. Department of Pediatrics I, Neonatology, University Children's Hospital Essen, University Duisburg-Essen, Essen, Germany
8. Child and Adolescent Psychiatry, Stockholm Health Care Services, Region Stockholm, Stockholm, Sweden
9. Curtin Autism Research Group, Curtin School of Allied Health, Curtin University, Perth, Western Australia
10. School of Education, Victoria University of Wellington, Wellington, New Zealand

* the authors contributed equally
~ Corresponding author

**ABSTRACT**

Video recording is a widely used method for documenting infant and child behaviours in research and clinical practice. Video data has rarely been shared due to ethical concerns of confidentiality, although the need of shared large-scaled datasets remains increasing. This demand is even more imperative when data-driven computer-based approaches are involved, such as screening tools to complement clinical assessments. To share data while abiding by privacy protection rules, a critical question arises whether efforts at data de-identification reduce data utility? We addressed this question by showcasing the Prechtl's general movements assessment (GMA), an established and globally practised video-based diagnostic tool in early infancy for detecting neurological deficits, such as cerebral palsy. To date, no shared expert-annotated large data repositories for infant movement analyses exist. Such datasets would massively benefit training and recalibration of human assessors and the development of computer-based approaches. In the current study, sequences from a prospective longitudinal infant cohort with a total of 19451 available general movements video snippets were randomly selected for human clinical reasoning and computer-based analysis. We demonstrated for the first time that pseudonymisation by face-blurring video recordings is a viable approach. The video redaction did not affect classification accuracy for either human assessors or computer vision methods, suggesting an adequate and easy-to-apply solution for sharing movement video data. We call for further explorations into efficient and privacy rule-conforming approaches for deidentifying video data in scientific and clinical fields beyond movement assessments. These approaches shall enable sharing and merging stand-alone video datasets into large data pools to advance science and public health.

**INTRODUCTION**

Data is the origin of knowledge gain and scientific progress. In recent decades, our private and professional lives have undergone rapid and significant changes in the way we receive, generate, and disseminate 'data'. Amongst the ever-expanding nearly uncontrollable sources and quantity of information in our daily life, rigorous data of scientific value are arguably rare. Obtaining high-quality data in behavioural science needs commitment of participants and the endowment of funders and institutions. Even more, elegant data owe a great debt to the ingenuity and unfailing efforts of the scientists who acquired them. When studying infant and child development, for example, data acquisition often requires longitudinal designs that can involve years or even decades of data collection to capture development across groups and settings. Video recordings are an important and widely used method of data collection in such studies. These data document dynamic human behaviours in real time and space. Sharing these valuable assets in the field, just like sharing data in other scientific fields, can maximise the benefits of resources, avoid redundant investment, improve study visibility, transparency and reproducibility, provide training resources, promote novel knowledge, and catalyse new cooperation [1–3].

Data sharing, however, faces a thicket of thorny issues, which have triggered contentious discourses by researchers, leading academic journals, funders and international organisations in the past years [4–12]. Although researchers are increasingly expected by funders, the scientific community, and tax payers to share data [6,13], a lack of actual incentives for primary researchers, aggravated by the extra operational costs and efforts necessary to curate data, impede such practice [14–16]. Especially when working with video data with identifiable individuals, ensuring the protection of participants' confidentiality is critical. Although major legal regulations such as the Health Insurance Portability and Accountability Act (HIPAA) of the United States and the General Data Protection Act (GDPR) of the European Union all obligate protection of data containing full-face images, which are regarded as sensitive personal identifiers, practical guidance on how such data can, if at all, be shared while protecting participants' confidentiality is scant [17,18]. For data collected in the past, for example, before GDPR was in force in 2018, clinicians and researchers could not foresee all the future needs and legal updates for data sharing beyond the scope of the original plan, hence may have missed the chance to obtain participants' consent for data sharing with other third parties (i.e., beyond what was consented to at the project start). How then, can we embrace societal interests and privacy frameworks and legitimate sharing of scientific video data? In this study, we addressed this issue by empirically testing a simple and widely used pseudonymisation approach in a specific scientific setting. Essentially, we would like to invite, through our work, fellow scientists and the community to revisit the discussion on data sharing, especially the sharing of video data, and seek practical approaches to make this happen, for the good of sustainable science and public health.

When investigating behavioural and neurofunctional development with infants, non-intrusive methods are especially desirable. Among such approaches, the *Prechtl's general movement assessment* (GMA;[19]) has become a worldwide established clinical tool, applicable during the very first months of human life, for identifying heightened risk for neurological impairment such as cerebral palsy [19–22]. GMA is renowned for its non-intrusiveness, excellent predictive validity, and peerless efficiency concerning required diagnostic time and resources ([23]; for methodological details please see [19,24]). The presence or absence of the fidgety movements (FM), for example, an age-specific motor pattern observable from the third to the fifth month in typically developing infants, has proven to be a highly sensitive and specific predictor for neurological deficits [20,21]. A standard GMA requires only a 3-minute guided video recording of an infant's spontaneous whole-body movements, which will be clinically evaluated by trained experts. It can be easily implemented in daily clinical routines or at the infant home, thus being especially flexible and suitable for either high- or low-resource settings. With GMA, high risk for neurological impairments can be excluded or detected [24], and potential intervention may be introduced early, which will mitigate long-term cost and burden for the health system [25]. GMA relies on human visual gestalt perception to classify typical vs. atypical infant motor patterns. As such, the excellence of the assessors does require specific high-quality training on the one hand, and continuous practice and recalibration on the other hand.

Despite the fact that GMA is accredited globally and indicated for application for the youngest population at risk for adverse neurological outcomes, the extent to which it has been scaled up in practice is still limited. One may presume that as artificial intelligence (AI) approaches can avoid unfavourable human and environmental factors affecting clinical reasoning, they are likely to have the potential to bolster GMA and outspread its application. Indeed, we have seen a boom of computer-based approaches to complement the classic man-powered GMA during the past decade [26]. Unfortunately, shared expert-annotated large GMA datasets are still absent. While large datasets are generally required to train machine learning algorithms, shared expert-annotated and approved large datasets are indispensable for evaluating and comparing performances of different AI approaches from different groups [27,28]. Moreover, as mentioned above, if such shared data repositories would be available within the scientific community, they would contribute enormously to train and recalibrate human GMA assessors. It is impossible for any single research or clinical site to accumulate sufficient amount of valid scientific data, for example through performing GMA, to cover diverse conditions of various aetiologies. To train human assessors, as well as computational models, to achieve reliable performance with high sensitivity and specificity, adequate data representing different classes (e.g., typical versus atypical GMs; movements from children with normal versus adverse neurological outcomes) are imperative. Sharing data across centres hence seems to be the ultimate way out.

Besides obtaining participants' informed consent to data sharing [29], algorithmic face blurring is a widely used approach for pseudonymisation. It protects the privacy of individuals when sharing visual data sources across industries from street mapping and social media to pictorial journalism. This technique has also been applied in the scientific fields such as neuroimaging and dentistry [30–33]. Face blurring commonly covers the eye region, which retains part of the facial expressions and could enhance data utility despite redaction. Face blurring is a straightforward method and is far easier to apply than, for example, generating avatars or synthetic surrogate faces [34,35]. Is it viable to apply face blurring approaches for video data sharing in research and practice in infant and child development? With the current study, we aim to examine the viability of using face blurring in this field by showcasing GMA, given its aforementioned scientific and clinical significance in child health.

In particular, we ask for the first time whether human assessors are able to perform comparable GMA when the infant faces are visible or blurred. Focusing on the classification of fidgety movements (presence vs. absence), we hypothesise that the performances of well-trained human GMA assessors do not differ in the two conditions: *Face-visible vs. Face-blurred*. Simultaneously, we ask whether AI methods could deliver comparable movement classifications using features with or without head key points (analogues of Face-visible vs. Face-blurred conditions). We hypothesise that the performances of the AI method are also comparable in the two different conditions.

**METHODS**

**Dataset and Participants**

To address our research questions, the baseline performance of human GMA assessors in the natural, face-visible condition is needed. These data are available and can be adopted from our previous study [27]. In that study, data from a prospective longitudinal cohort of 51 typically developing infants were analysed [27]. Data acquisition was conducted at iDN's BRAIN*tegrity* lab at the Medical University of Graz, Austria, within an umbrella study profiling typical cross-domain development during the first months of life[36]. The movement data in form of RGB video stream were collected in a standard laboratory setting following the Prechtl's general movements Assessment guidelines [24]. Details on data recording and participants information have been described before [27,36,37]. For the machine learning algorithm presented in that study, 2800 five-second video-snippets were randomly selected from a total of 19,451 available snippets. This dataset of 2800 snippets was used for the current study. Data segmentation, annotation and analyses were performed at the Systemic Ethology and Developmental Science Unit - SEE, Department of Child and Adolescent Psychiatry and Psychotherapy at the University Medical Center Göttingen, Germany. The study was approved by the Institutional Review Board of the Medical University of Graz, Austria (27-476ex14/15) and the University Medical Center Göttingen, Germany (20/9/19). Parents were informed of all experimental procedures and study purpose, and provided their written informed consent for participation and publication of results.

**Movements classification by human assessors in the *face-visible* condition**

In our previous study, the 2800 snippets were annotated by two well-trained and experienced human GMA assessors. The infant faces were visible to the assessors, as in a standard general movements assessment. The assessors, independent from each other, classified each snippet as "fidgety movements present" (FM+), "fidgety movements absent" (FM-), or "not assessable" (i.e., the infant presents one or more of the following states during the specific 5 s: fussy/crying, drowsy, yawning, refluxing, over-excited, self-soothing, or distracted by the environment, all of which distort infants' movement pattern and shall not be assessed for GMA, [24]). Out of the 2800 snippets, 990 were labelled by at least one assessor as "not assessable", mirroring infants' frequently fluctuating behavioural status.

Of the remaining 1810 snippets, 1784 snippets (98.6%) were labelled identically by both assessors: Either FM+ (N = 956) or FM- (N = 828). For classes FM+ and FM-, the interrater agreement was excellent (Cohen's kappa κ = 0.97, with .95 CI [.96, .98]). The intra-rater reliability by rerating 280 randomly-chosen snippets (i.e. 10% of the sample) for the two classes was Cohen's kappa κ = .95 with .95 CI [.91, 1] for assessor 1, and κ = 0.85 with .95 CI [.78, .93] for assessor 2. These data are adopted by the current study as the classification accuracy for human assessors in the face-visible condition. To conduct new experiments in the face-blurred condition, we carried out a face masking procedure with the original 2800 snippets.

**Face blurring procedure**

A flow diagram of the face blurring procedure is shown in Fig. 2 and consists of two steps: (1) extraction of body key points (see Fig. 1) in order to determine position of eyes and nose (Fig. 2(b)), and (2) face masking by applying blurring filter in the area around eyes and nose (Fig.2(c)).

***Extraction of body key points.*** To determine the position of the mask, i.e., the anatomical area around the eyes and the nose, we used a state-of-the-art pose estimation method OpenPose [38]. OpenPose is based on deep learning and extracts 25 body key points from 2D images including five head key points, i.e., eyes, nose, and ears (see Fig. 1). OpenPose was already successfully used in recent studies on classification of infant movements [27,39,40]. In this study, we also used OpenPose body key points not only to determine the position of the mask but also as features for movement classification and analysis about the importance of head key points.

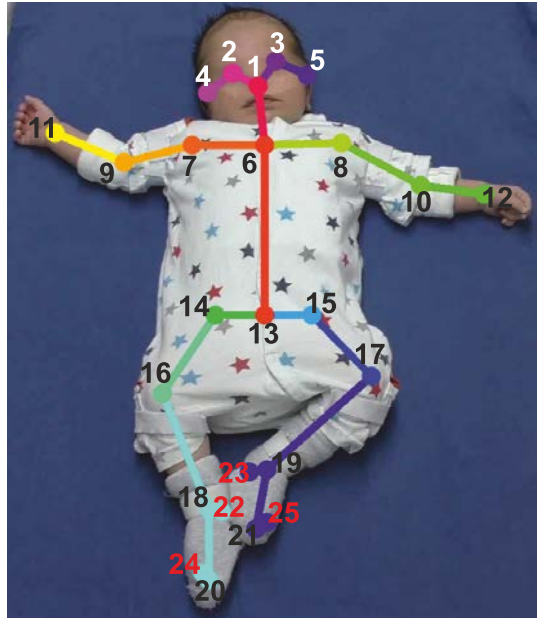

**Figure 1:** Example of extraction of body key points using OpenPose [38]. We used key points 1-5 for face blurring, and key points 1-21 (with head key points) or key points 1-16 (without head key points) for movement classification. Key points 22-25 were not used in this study due to poor position estimation of these key points.

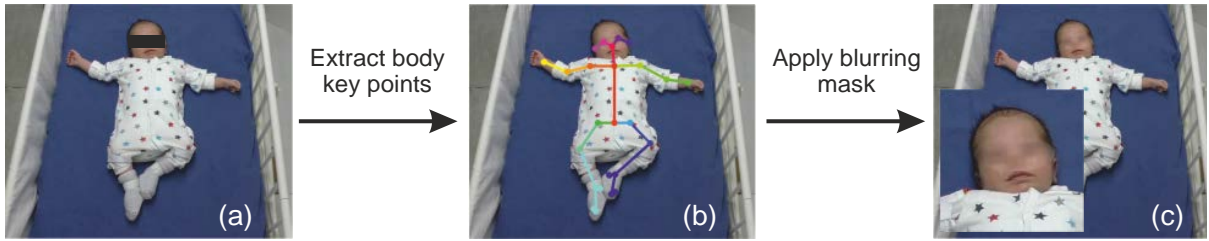

**Figure 2:** Flow diagram of face blurring procedure: (a) original image, (b) extracted body key points using OpenPose [38], (c) obscured face with a blurring mask.

For each frame in the video (5s x 50fps = 250 frames with resolution of 1920 x 1080 pixels) we extracted 25 body key points (see Fig. 1). For each body key point, OpenPose returns three values: $x\_i$ and $y\_i$ coordinates in the image and a reliability score $r\_i$ between 0 and 1 of each key point ($i$ = 1 … 25). The reliability score defines how confident the algorithm is in predicting coordinates of the key point. Five head key points 1-5 (eyes, nose, and ears) were used to determine the position of the mask, whereas key points 1-21 (excluding toes and heels) or key points 6-21 (excluding toes, heels and five head key points) were used for movement classification.

***Face blurring.*** As stated above, we used five head key points 1-5 (nose, eyes and ears; see Fig. 1) to determine the position of the mask for face blurring. A centre point coordinates $c\_x$ and $c\_y$ of the elliptic mask was defined as average values of head key points:

$c\_x(f)$ = mean([$eyL\_x, eyR\_x, ns\_x, erL\_x, erR\_x$]),

$c\_y(f)$ = mean([$eyL\_y, eyR\_y, ns\_y$]),

if average reliability score $r\_avg$ = mean([$eyL\_r, eyR\_r, ns\_r$]) > 0.35, otherwise

$c\_x(f) = c\_x(f-1)$,

$c\_y(f) = c\_y(f-1)$,

where $f = 1 \ldots 250$ denotes the frame number, *eyL* – left eye, *eyR* – right eye, *ns* – nose, *erL* – left ear, *erR* – right ear. Note that for *c_y* we only used eye and nose key points since we wanted to mask the area around eyes and nose but leave the mouth area still visible (please see Discussion).

We have also applied an exponential moving average filter to reduce jerk of the mask movement in video:

$c\_x(f) = a*c\_x(f-1) + (1-a)*c\_x(f)$,

$c\_y(f) = a*c\_y(f-1) + (1-a)*c\_y(f)$, with $a = 0.5$.

Finally, we applied an elliptic blurring mask with the centre point c_x, c_y, width $w = 150$ pixels and height $h = 68$ pixels. To generate the blurring mask, we used the normalized box filter using standard OpenCV [OpenCv] blurring function with kernel size $k = 25$. In addition, to prevent reconstruction we also applied random noise to the blurred pixels from uniform distribution between 0 and 25 (which on average corresponds to about 5% of relative noise).

**Movement classification by human assessors in the *face-blurred* condition**

From the 2800 snippets, three subsets of 280 snippets each were randomly selected. Snippets in the subsets did not overlap with each other. The two experienced GMA assessors from the previous study [27] independently rated the three subsets of face-blurred snippets, making the FM+, FM-, or not assessable classifications. The current rating took place 8 months after the original assessments, so that the memory effect is barely conceivable. The presenting order of the three subsets were the same for both assessors. The assessors did not receive feedback, nor did they communicate with each other throughout the assessment. The rating procedure remained identical as in the previous study, except that the assessors only saw infants with blurred faces. The classification results of the three subsets will be compared to the assessors' original classifications (i.e., face-visible condition) summarised above.

**Movement classifications for face-blurred and face-visible conditions with a machine learning approach**

After applying the face blurring mask, the head key points could not be detected reliably. Thus, the question arises whether head key points are indispensable features for the specific fidgety movement classification; i.e., whether classification accuracy would decrease significantly when head key points are excluded. To answer this question, we performed classification of fidgety movements into two classes (FM+ and FM-) when using features with and without head key points. Note that the goal here is not to arrive at the network architecture with the highest classification accuracy but to compare classification performances in the two conditions.

*Features.* We used OpenPose body key points (x and y coordinates) extracted from all video frames (250 in total) as features for a neural classifier (see below). As addressed above, we used 21 key points (without toes and heels) or 16 key points (without toes and heels and *without* five head key points). Note that for classification experiments we extracted body key points using the original videos. We discarded toes and heels because these key points were not detected reliably using OpenPose. Thus, we constructed a feature matrix of size 250x42 or 250x32 depending on whether we used head features or not.

In some cases, e.g., due to occlusions, OpenPose cannot detect coordinates of key points in images and these values by default are set to 0. Since we operated with videos and not with single images, we set 0 values to the values obtained from linear interpolation between frames with non 0 values. In addition, we applied min-max normalisation (normalised values between 0 and 1) to remove influence of body size and subtracted average value calculated across frames for respective coordinates values of each key point.

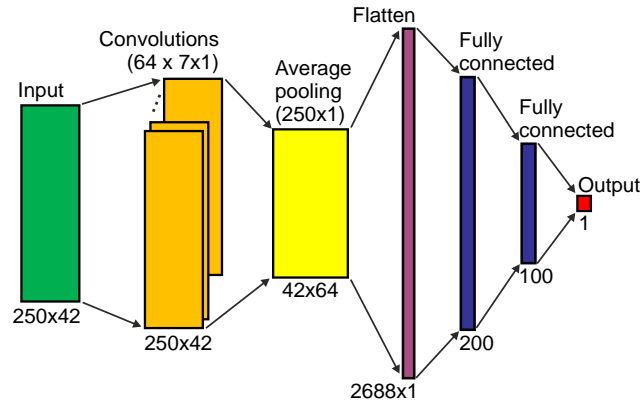

**Figure 3:** Network architecture with one convolutional layer with 64 filters of size 7x1, and two fully connected layers with 200 and 100 neurons, respectively. 250 corresponds to the number of video frames (5s x 50 frames/s) and 42 corresponds to the number of features with head key points (21 key points with x and y coordinates for each key point). We also used a batch normalisation and a dropout of 10% after convolutional and fully connected layers (not shown).

***Network architectures and training procedure.*** We used simple shallow multi-layer network architectures with one convolutional layer and one or two fully connected layers (see Fig. 3). Again, our goal here is not to arrive at the network architecture with the highest classification accuracy but to compare performances when using features with and without head key points. For more sophisticated network architectures see [27,40]. We also performed an ablation study where we investigated network architectures with different number of neurons in the fully connected layers and different number of filters and different filter sizes in the convolutional layer (see Table 2 and Table 3).

For network training we used the Adam optimiser with the binary cross-entropy as a loss function and the batch size of 32 samples. To prevent the network from overfitting we used validation stop (1/8 of training data) where we stopped training if classification accuracy on the validation set was not improving in ten consecutive epochs. For each parameter set we performed training of the network ten times and then selected the model with the classification accuracy on the validation set which then was evaluated on the test set. Neural networks were implemented using TensorFlow (https://www.tensorflow.org/) and Keras API (https://keras.io/).

***Evaluation procedure.*** As presented above, our dataset consisted of 1784 samples (956 FM+ and 828 FM-). For comparison of movement classification performance when using models with and without head features, we performed a 5-fold cross-validation, where each time 1/5 of the data was used for testing and the rest 4/5 of the data was used for training. Note that training data was split into training set (7/8 of the data) for parameter update and validation set (1/8 of the data) for training stop (see above). We used mean classification accuracy obtained from five test sets to analyse importance of head features for classification performance.

## RESULTS

**Movement classification by human assessors**

*Comparing each assessor's performance in the face-blurred condition to that in the face-visible condition*

The performances of the two assessors in the current face-blurred condition are compared with their own ratings in the original face-visible condition. Classification results for the classes FM+ vs. FM- are presented in Table 1 for the two conditions for each assessor. For both assessors and across all the three subsets, the classifications (FM+ vs. FM-) in the face-blurred condition presented excellent to perfect agreements with each assessor's original classification in the face-visible condition. Recall that in the original face-visible condition, the intra-rater agreement of a subset of randomly chosen 280 snippets was .95 (.95 CI [.91, 1]) for assessor 1, and .85 (.95 CI [.78, .93]) for assessor 2. In other words, the assessors' performance were not affected by the factor whether the infant faces were visible. Note that the assessors did not receive any feedback nor communicate with each other throughout the rating procedure. Still, for both assessors, the performances, i.e., the agreement between the face-blurred vs. face-visible conditions, improved from the first to the third subset.

**Table 1.** Intra-rater agreement (Cohen's kappa κ and [.95 CI]) for classes FM+ vs. FM- between the original face-visible condition and the face-blurred condition for the three non-overlapping randomly-chosen data subsets each of 280 snippets.

|  | Subset 1 | Subset 2 | Subset 3 | **Combined** (840 snippets) |
|---|---|---|---|---|
| **Assessor 1** | .90 [.83, .97] | .97 [.93, 1] | .97 [.94, 1] | .95 [.92, .97] |
| **Assessor 2** | .87 [.79, .94] | .93 [.88, .99] | .94 [.89, .99] | .91 [.88, .95] |

*Inter-rater agreement between assessors in the face-blurred condition*

In the original face-visible condition, the two assessors' agreement with each other for classes FM+ and FM- with the entire sample (i.e., 2800 snippets) was Cohen's kappa κ = .97 with .95 CI [.96, .98]. In the current experiment with face-blurred snippets, the agreement on classes FM+ and FM- between the two assessors was .78 with .95 CI [.69, .87] for the first, .89 with .95 CI [.82, .96] for the second, and .99 with .95 CI [.99, 1.00] for the third subset of snippets. Again, although there was no feedback to the assessors nor communication between the assessors all along, the inter-rater agreements between the assessors also increased from the first to the third subset.

*Class "not assessable"*

As the assessors in the face-blurred condition can only see obscured faces, judging whether or not a 5-second snippet was "not assessable" (i.e., the infant presents one or more of the following states: fussy/crying, drowsy, yawning, refluxing, over-excited, self-soothing, or distracted by the environment) was challenging. Obviously, if the assessors could see the entire faces of the infants, they could more easily identify the non-assessable snippets. Out of the 840 snippets, more were labelled as "not assessable" in the face-visible condition (277 by assessor 1, and 249 by assessor 2) than were in the face-blurred condition (266 by assessor 1, and 188 by assessor 2). For assessor 1, 184 out of the 840 snippets were labelled in both conditions as "not assessable", and for assessor 2, 123 were.

**Movement classification with machine learning approach**

Results of classification performance when using networks with and without head key points are presented in Tables 2 and 3. Classification accuracies in most of the cases are above 86% and below 88% for both models with or without head features. In Fig. 4 we compare best classification accuracy scores obtained for the networks with one and two fully connected (FC) layers. When comparing network architectures, results show that although on average networks with two FC layers lead to better classification performance than networks with one FC layer (**85.03%** and **86.94%** vs. **87.16%** and **88.17%, see Fig. 4**), this difference is not significant (two-sample t-test, p>0.05). The current classification accuracy is comparable to our previous study on FM classification [27] and to a more sophisticated neural architecture of [40].

Most importantly, when comparing models with and without head features, results demonstrate that there is no statistically significant difference when comparing classification accuracy with and without head key points: p = 0.088 and p = 0.6368 for the network with one FC layer and with two FC layers, respectively (two-sample t-test). Results show that the absence of head key points does not have a significant effect on the specific classification of fidgety movements (i.e., FM+ vs. FM-). This suggests that the proposed face blurring approach, in contrast to the one without face blurring, can deliver comparable classification results for fidgety movements when using body pose as features without head key points.

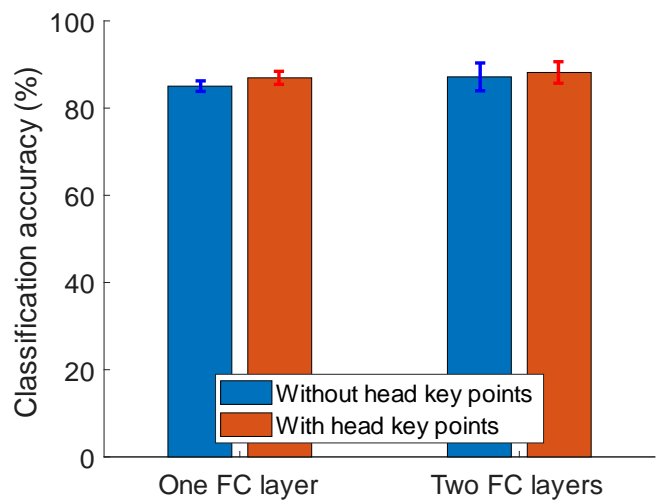

**Figure 4:** Classification accuracy of fidgety movements without and with head key points when using network architectures with one and two fully connected (FC) layers, respectively. Error bars denote confidence intervals of mean (95%). Network parameters were as follows. **One FC layer without head key points:** 64 filters of size 7x1 in the convolutional layer, and 300 neurons in the fully connected layer; **one FC layer with head key points:** 64 filters of size 7x1 in the convolutional layer, and 100 neurons in the fully connected layer; **two FC layers without head key points:** 64 filters of size 7x1 in the convolutional layer, and 300 and 200 neurons in the fully connected layers; **two FC layers with head key points:** 64 filters of size 31x1 in the convolutional layer, and 500 and 250 neurons in the fully connected layers.

**Table 2:** Classification accuracy of fidgety movements without and with head key points when using network architectures with one and two fully connected layers, respectively. Bold Numbers represent highest average classification accuracy within each group. Error bars denote confidence intervals of mean (95%).

|  | **Without head key points** |  | **With head key points** |  |
|---|---|---|---|---|
| Number of neurons | **One fully connected layer** (64 filters of size 7x1) | | | |
|  | Mean | CI (95%) | Mean | CI (95%) |
| 50 | 83.5219 | ±2.1751 | 86.8297 | ±2.5079 |
| 100 | 83.4665 | ±2.6224 | **86.9402** | ±1.4941 |
| 150 | 84.0259 | ±1.5520 | 86.8838 | ±0.4613 |
| 200 | 84.2511 | ±2.2011 | 85.8183 | ±1.8428 |
| 300 | **85.0349** | ±1.2096 | 86.1557 | ±1.4715 |
| 500 | 83.8589 | ±3.0408 | 86.7150 | ±1.2738 |
| Number of neurons | **Two fully connected layers** (64 filters of size 7x1) | | | |
|  | Mean | CI (95%) | Mean | CI (95%) |
| 50, 25 | 84.3616 | ±2.7607 | 86.4363 | ±1.9307 |
| 100, 50 | 85.1465 | ±1.3817 | 86.6028 | ±0.5126 |
| 150, 100 | 86.7712 | ±2.0580 | 87.1074 | ±0.8705 |
| 200, 100 | 86.0432 | ±2.7556 | 86.4914 | ±1.6482 |
| 300, 150 | 86.0435 | ±2.3402 | 84.5860 | ±1.5265 |
| 300, 200 | **87.1636** | ±3.1892 | 87.1628 | ±1.3906 |
| 500, 250 | 84.4717 | ±2.4624 | **87.2206** | ±1.6098 |
| 500, 300 | 85.4268 | ±1.2196 | 85.9311 | ±0.9800 |

**Table 3:** Classification accuracy of fidgety movements without and with head key points when using different filter sizes and different numbers of filters in the convolutional layer, respectively. The network with two fully connected layers was used in this case with 200 and 100 neurons per layer. Bold numbers represent highest average classification accuracy within each group. Error bars denote confidence intervals of mean (95%).

|  | **Without head key points** |  | **With head key points** |  |
|---|---|---|---|---|
| **Filter size** | 64 filters, two fully connected layers (200, 100) | | | |
|  | Mean | CI (95%) | Mean | CI (95%) |
| 5x1 | 72.6364 | ±10.0515 | 76.6857 | ±6.3535 |
| 7x1 | 86.0432 | ±2.7556 | 86.4914 | ±1.6482 |
| 9x1 | **86.2668** | ±1.8780 | 86.9394 | ±1.2111 |
| 15x1 | 84.0270 | ±3.5293 | 87.5558 | ±1.4402 |
| 21x1 | 84.7555 | ±1.8379 | 87.8927 | ±0.5532 |
| 31x1 | 85.8750 | ±2.0184 | **88.1733** | ±2.4673 |
| **Number of filters** | Filter size 7x1, two fully connected layers (200, 100) | | | |
|  | Mean | CI (95%) | Mean | CI (95%) |
| 16 | 85.5948 | ±2.0099 | 84.6407 | ±1.1507 |
| 32 | 84.6442 | ±3.9150 | 85.2024 | ±1.9710 |
| 64 | 86.0432 | ±2.7556 | 86.4914 | ±1.6482 |
| 128 | 86.2682 | ±2.1049 | 85.9864 | ±1.5832 |
| 256 | **86.5480** | ±3.0186 | **86.7161** | ±2.0170 |
| 512 | 84.8097 | ±1.9802 | 85.9873 | ±1.4872 |

## DISCUSSION

Despite increasing expectations from leading funding organizations and journals for data-sharing, it appears to be more desired than practised [41–44]. Even many researchers declare their willingness to share data behind the published articles, most do not respond or decline data access requests when asked [41]. The strikingly low reported compliance rate (i.e., less than 7%) was the same as for authors who did not provide a willing-to-share-data statement [41]. Barriers seem to persist to stop scientists actually sharing their data [45,46]. Many researchers decline data sharing because of ethical issues, including ensuring confidentiality or lacking informed consent for sharing unless participants can be deidentified. This is even more true when it comes to video data sharing, where participants' identities are likely to be more difficult to conceal, and little practical guidance is available to support video data sharing.

While de-identification is certainly a key issue in video data sharing to protect participants' confidentiality, the question arises naturally whether efforts at data redaction reduce data utility. Our current study explored the feasibility of one simple and widely-used approach, i.e., face-blurring, for pseudonymisation of video data in a scientific setting assessing infant movements. Specifically, we tested whether classification performances of both human GMA assessors and machine learning approach remain sufficient after videos are pseudonymised.

Our data suggest that well-trained experienced human GMA assessors' performances are not negatively affected by face-blurring, although the assessors may benefit from a brief adaptation to the altered video presentation in practice (i.e., rating face-obscured infants). This was suggested by the slightly lower, although still excellent accuracy of classification for the very first subset of the experiment, by both assessors, followed by even higher, nearly perfect performance in the later subsets fully comparable to that in the standard face-visible condition (Table 1). Note that the assessors in this study rated the face-blurred snippets without getting familiarised to the video presentation (face-blurred) beforehand, thus the result exemplifies the true performance variations, if any, that human GMA assessors may experience with this specific pseudonymisation approach. Future studies need to sample more assessors with diverse GMA experiences to examine whether different raters are able to work with face-blurred GMA video data and deliver unaffected movement classification.

For the presented machine learning approach, no significant difference in classification performance comparing models including or excluding head points (analogues for the Face-visible vs. Face-blurred conditions) was found. As stressed before, our goal was not to arrive at a network architecture with the highest classification accuracy but to detect whether classification performances differ in these two conditions. Our data verified that, for detecting the presence or absence of an age-specific infant motor pattern (i.e., fidgety movements), the AI approach works equally well with deidentified, facial-feature-excluded video data.

Our results suggest the viability, for both human assessors and computer-based methods, of using face-blurring techniques to pseudonymise and share video data for movement analyses. These are promising news for scientists worldwide who study infant neuromotor development. Movements video data can be deidentified without reducing utility, enabling multicentred sharing and pooling data. In light of the GDPR, the challenge of proper application of pseudonymisation to personal data remains. Indeed, there is no single easy solution that works for all research purposes in all possible scenarios [17]. Moreover, wherever there is a pseudonymisation technology, there might be an opposing approach (i.e., reidentification) to it [47]. The authors are aware that the technical solution discussed in this study may not be applicable for another research setting, such as studies evaluating footages of social interaction, where participants' facial images cannot be removed for the sake of data sharing which would consequently eliminate data utility. With our study, we intend to promote exploration and discussion on efficient and innovative solutions to sharing different types of valuable video data documenting human behaviours in clinical and scientific settings. This includes solutions to existing data for which participants' consent for data sharing may not be available or achievable any more. Some technical attempts to deidentify individuals on video footages while preserving their dynamic facial attributes in real-time have been made [34,48,49]. Such approaches need to be empirically tested for their utility, reliability, and efficiency for easy implementation in research and clinical practices.

Data sharing is a systemic ambition requiring diverse skillsets encompassing scientific, technological, financial, administrative, political, ethical and legal issues [10,11,50–52]. Scientists who welcome data sharing calls and regulations urgently need systematic in-practice support from policy makers, from professionals and experts in data curation and data protection, so that their capacity and mindset may focus mainly on the subject matter of science. Individual video data may belong to the most sensitive and resource-consuming type of data to collect, curate, and process, within and beyond infant and child development research. If researchers do not have to worry that their data might be misinterpreted or misused by other beneficiaries, or their original ideas behind the data might be scooped, more scientist might embrace and practise data sharing and exchange. Scientists who share video data behind their studies ought to be fairly and adequately accredited and rewarded for their investment. This way, data sharing, especially video data sharing will become incentivised and fruitful, connecting disciplines in scientific and clinical communities, ultimately benefitting science and public health.

**ACKNOWLEDGEMENTS**

First and foremost, we would like to thank all families for their participation, their support for basic science, their patience with scientists and commitment to our studies. The authors would like to thank team members involved in recruitment, data acquisition and curation: Magdalena Krieber-Tomantschger, Iris Tomantschger, Laura Langmann, Dr. Robert Peharz, Dr. Florian Pokorny, Claudia Zitta and Gunter Vogrinec. We thank Lennart Jahn and Dr. Simon Reich for data pre-processing. We were/are supported by BioTechMed Graz and the Deutsche Forschungsgemeinschaft (DFG – stand-alone grant, SFB1528), the Laerdal Foundation, the Bill and Melinda Gates Foundation (OPP1128871), the Volkswagenfoundation (project IDENTIFIED), the LeibnizScience Campus, the BMBF Germany, and the Austrian Science Fund (KLI811), and the Fondation Paralysie Cerebrale (ENSEMBLE-II) for data acquisition preparation and analyses. Special thanks also to our interdisciplinary international network of collaborators for discussing this study with us and for refining our ideas and analytical processes in a global consortia approach.